\documentclass[letterpaper]{article} % DO NOT CHANGE THIS
\usepackage{aaai23}  % DO NOT CHANGE THIS
\usepackage{times}  % DO NOT CHANGE THIS
\usepackage{helvet}  % DO NOT CHANGE THIS
\usepackage{courier}  % DO NOT CHANGE THIS
\usepackage[hyphens]{url}  % DO NOT CHANGE THIS
\usepackage{graphicx} % DO NOT CHANGE THIS
\usepackage{natbib}  % DO NOT CHANGE THIS AND DO NOT ADD ANY OPTIONS TO IT
\usepackage{caption} % DO NOT CHANGE THIS AND DO NOT ADD ANY OPTIONS TO IT
\usepackage{algorithm}
\usepackage{algorithmic}

\usepackage{newfloat}
\usepackage{listings}
\DeclareCaptionStyle{ruled}{labelfont=normalfont,labelsep=colon,strut=off} % DO NOT CHANGE THIS
\floatstyle{ruled}
\newfloat{listing}{tb}{lst}{}
\floatname{listing}{Listing}
\title{Denoising Multi-Similarity Formulation: A Self-paced  Curriculum-Driven Approach for Robust Metric Learning}
\author {
    Chenkang Zhang\textsuperscript{\rm 1},
    Lei Luo\textsuperscript{\rm 2}, 
    Bin Gu \textsuperscript{\rm 1,\rm 3}\thanks{Corresponding Author}
}
\affiliations {
    \textsuperscript{\rm 1} School of Computer and Software, Nanjing University of Information Science and Technology, P.R.China\\
    \textsuperscript{\rm 2} School of Computer Science and Engineering, Nanjing University of Science and Technology, P.R.China\\
    \textsuperscript{\rm 3} MBZUAI, United Arab Emirates\\
    20201221058@nuist.edu.cn, luoleipitt@gmail.com, jsgubin@gmail.com
}

\usepackage{bibentry}
\usepackage{amsmath}
\usepackage{amssymb}
\usepackage{mathtools}
\usepackage{amsthm}

\usepackage{multirow}
\usepackage{bbding}
\usepackage{amsmath}\allowdisplaybreaks 
\usepackage{graphicx} 
\usepackage{algorithm}
\usepackage{algorithmic}
\usepackage{setspace}
\usepackage{bm,enumitem,caption}
\usepackage{subfigure}
\usepackage{wrapfig}
\usepackage{multirow}
\newsavebox\CBox

\newtheorem{theorem}{Theorem}
 
\newtheorem{remark}{Remark}
\newtheorem{assumption}{Assumption}
\newtheorem{definition}{Definition}

\DeclareMathOperator{\e}{\mathbf{e}}
\DeclareMathOperator{\w}{\mathbf{w}}
\DeclareMathOperator{\x}{\mathbf{x}}

\begin{document}

\maketitle

\begin{abstract}
Deep Metric Learning (DML) is a group of techniques that aim to measure the similarity between objects through the neural network. Although the number of DML methods has rapidly increased in recent years,  most previous studies cannot effectively handle noisy data, which commonly exists in practical applications and often leads to serious performance deterioration. To overcome this limitation, in this paper, we build a connection between noisy samples and hard samples in the framework of self-paced learning, and propose a \underline{B}alanced \underline{S}elf-\underline{P}aced \underline{M}etric \underline{L}earning (BSPML) algorithm with a denoising multi-similarity formulation, where noisy samples are treated as extremely hard samples and adaptively excluded from the model training by  sample weighting. Especially, due to the pairwise relationship  and a new balance regularization term, the sub-problem  \emph{w.r.t.}  sample weights is a  nonconvex  quadratic function.  To efficiently solve this nonconvex  quadratic problem, we propose a doubly stochastic projection coordinate gradient algorithm. Importantly, we  theoretically prove  the convergence  not only for the doubly stochastic projection coordinate gradient algorithm, but also for our BSPML algorithm. Experimental results on several standard  data sets demonstrate that our BSPML algorithm has   better generalization ability and robustness  than  the state-of-the-art   robust DML approaches.
\end{abstract}

\section{Introduction}
DML aims to learn an embedding space  in which similar samples are pulled closer while dissimilar samples are encouraged to stay away from each other \citep{xing2002distance,wang2021robust,wang2022metricmask}. Compared with traditional metric learning methods, which may not capture the nonlinear nature of data, DML utilizes the neural network to obtain   representative and discriminative feature embeddings. Thus, DML has attracted increasing attention and been applied to various tasks, including  visual tracking \citep{leal2016learning,tao2016siamese}, face recognition \citep{wen2016discriminative}, image retrieval \citep{wohlhart2015learning,he2018triplet,grabner20183d},   person re-identification \citep{hermans2017defense,yu2018hard} and zero-shot learning \citep{zhang2016zero,yelamarthi2018zero,bucher2016improving}.

The superior performance of machine learning greatly depends on a large number of labeled data sets \citep{shi2021triply}, and the performance of DML is no exception. However, manually generating clean data set would involve domain experts evaluating the quality of collected data and thus is very expensive and time-consuming \citep{frenay2013classification}. To address this issue, some researchers utilize the online key search engine method \citep{yu2018learning} and the crowdsourcing method \citep{li2017webvision}  to gain required data sets at a  low cost, but it is possible to introduce \textbf{noisy samples} that represent mislabeled ones \citep{wu2021ngc,yao2021jo,zhai2020safe}. As far as we know,  most existing DML approaches are sensitive to noisy samples since they directly utilize sample labels to learn the similarity information between samples.

Constructing  robust DML models  against noisy samples is  a challenging task, and
some researchers have paid attention to this problem. Specifically,  Wang \emph{et al.}   proposed a novel objective using the $\ell_{1}$-norm distance \citep{l1norm}, and  Al-Obaidi \emph{et al.}  utilized the  rescaled hinge loss, which was a general form of the common hinge loss,  to formulate the DML problem \citep{rescaled}. Moreover,  Kim \emph{et al.} proposed the  Proxy-Anchor loss that  was robust against noisy labels and outliers because of the use of proxies \citep{proxyanchor}.  Different from the above technical routes, Yuan \emph{et al.}  proposed a robust distance metric based on the Signal-to-Noise Ratio (SNR) \citep{yuan2019signal}.

Different from mislabeled noisy samples, \textbf{hard samples} are correctly labeled ones that are difficult to distinguish by the model \citep{nguyen2019self,zhu2021hard}. In the image data set, hard samples can be anything from cats that look like dogs to images with slightly blurred resolution. Evidently,  noisy samples and hard samples have different roles. Noisy samples are harmful because they would mislead the training direction. However, hard samples could force the model to learn more representative features, and thus properly training these hard samples could improve the model generalization ability \citep{zhu2021hard,chen2020hard}. Although noisy samples and hard samples are different, we would build the connection between them in the perspective of  Self-Paced Learning (SPL) \citep{kumar2010self}, and further utilize SPL to filter noisy samples out from DML.

\begin{figure}[t] 
	\centering
	\includegraphics[width=0.45\textwidth]{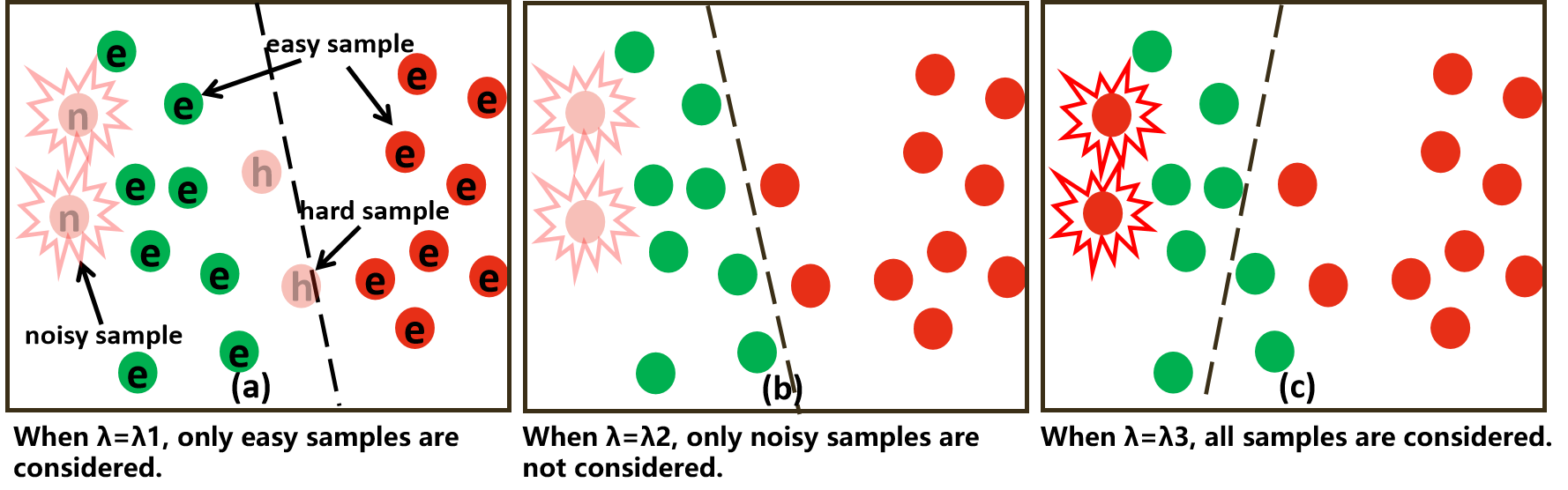}
	\vspace{-4pt}
	\caption{Self-paced classification problem with  hard  and  noisy samples.  ($\lambda$ is the age parameter and $\lambda_1 < \lambda_2 < \lambda_3$.)} \label{SPL}
	\vspace{-8pt}
\end{figure}

Inspired by human cognitive mechanisms, \citep{kumar2010self} proposed a novel learning strategy called Self-Paced Learning (SPL), which utilizes the loss values to describe the difficulties of samples. Also, SPL starts learning from easy samples and gradually incorporate more difficult samples (\emph{i.e.}, hard samples). Particularly, we could consider the noisy samples as the extremely hard samples because they usually own  
larger loss values than conventional hard samples (please refer to Fig.  \ref{SPL}.(a)). Thus,  SPL can enforce smaller weights to noisy samples under the guidance of loss values, and then noisy samples would have less influence on the predicted model. 
Theoretically, \citep{meng2017theoretical} has proved that such a re-weighting learning process is equivalent to minimizing a latent noise-robust loss that would weaken the contribution of noisy samples.
Moreover, it has been reported that SPL is an effective method to improve the robustness of the model against noisy data  \citep{zhang2020self,yin2021self,ren2020self,gu2021finding}.

In this paper, we propose a \underline{B}alanced \underline{S}elf-\underline{P}aced \underline{M}etric \underline{L}earning (BSPML) algorithm with a novel denoising multi-similarity formulation.   Benefiting from the mechanism of SPL, our BSPML algorithm could exclude noisy samples and emphasize the importance of clean samples.
Because DML problems focus on a large number of classes, we have to face the challenge of the unbalanced average sample weights among classes. To overcome this problem,  we introduce a balance regularization term to punish the absolute difference between the average sample weights of different classes. Following the traditional SPL practice, our BSPML algorithm utilizes the alternative optimization strategy based on two key sub-problems  \emph{w.r.t.}   model parameters and sample weights respectively. To efficiently solve the nonconvex sub-problem \emph{w.r.t.}  sample weights, we propose a doubly stochastic projection coordinate gradient algorithm. Theoretically, we prove the convergence of our proposed algorithms under mild assumptions. Experimental results show the advantages of our BSPML algorithm in generalization ability and robustness.

\begin{figure}[t] 
	\centering
	\includegraphics[width=0.4\textwidth]{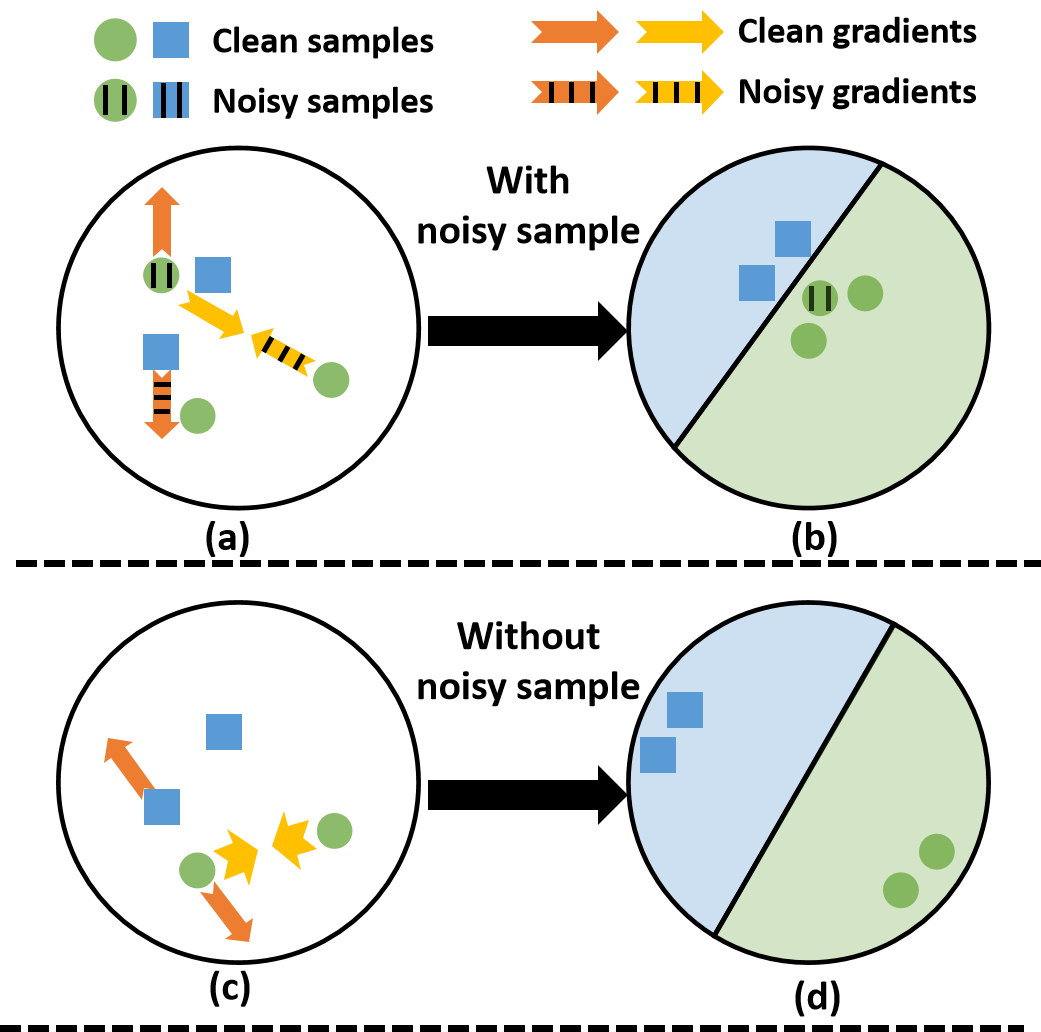}
		\vspace{-4pt}
	\caption{The difference between the two-class DML tasks with and without the noisy sample. (A pair of pointers of the same color in one sub-figure represents  a pair of gradients.)} \label{withoutSPL}
		\vspace{-8pt}
\end{figure}

\section{Preliminaries}
In this section, we give a brief review of self-paced learning and multi-similarity loss.

\subsection{Self-paced learning}

Suppose $\{(\x_i,y_i)_{i=1}^N\}$ is a set of $N$ labeled samples, where $y_i\in [C]$ is  the  corresponding  label  of the sample $\x_i \in  \mathbb{R}^M$.  SPL utilizes a sample weight vector $\w \in [0,1]^N$ to indicate whether or not each training sample should be included in the current training. The classic SPL model  is formed as:
\begin{align} \label{EqclassicSPL}
    \min\limits_{\theta,\w \in [0,1]^N } &\sum_{i=1}^N w_i L_{\text{point}}(\x_i,y_i,\theta) - \lambda \sum_{i=1}^N w_i, 
\end{align}
where $\theta$ means the model parameters, $L_{\text{point}}$ represents one point-wise loss, \emph{e.g.}, hinge loss, $\lambda$ is the age parameter which controls the learning pace in SPL. 

SPL methods utilize the alternative  optimization strategy based on two key sub-problems \emph{w.r.t.}  model parameters $\theta$ and sample weights $\w$ respectively.  Especially, for the classic SPL formulation, the closed-form solution of the sub-problem \emph{w.r.t. } $\w$  can be  obtained easily:
\begin{equation} \label{Eqeasy}
\left\{
    \begin{aligned}
    &w_i=1, \ \   L_{\text{point}}(\x_i,y_i,\theta) \leq \lambda  \\
    &w_i=0, \ \    L_{\text{point}}(\x_i,y_i,\theta)  > \lambda
    \end{aligned}
\right . .
\end{equation}
The above solution implies that if the loss value of a sample is less than $\lambda$, the sample will be assigned the weight of value $1$ and thus be selected to join the training. Otherwise, the sample will be excluded from the training process by reducing its weight to 0. In this case,  with increased $\lambda$, hard samples with larger loss values will join the training. Thus,  the age parameter  $\lambda$ controls the learning pace in SPL.

To better understand the mechanism of SPL and the roles of hard and noisy samples, we provide an example of the self-paced classification problem in Fig. \ref{SPL}. As shown in  Fig. \ref{SPL}.(a),  hard samples have larger loss values than easy samples, and noisy samples can be considered as extremely hard samples. According to Eq. (\ref{Eqeasy}), when we set $\lambda$ to a small value $\lambda_1$,  only easy samples are selected to join the training in Fig.  \ref{SPL}.(a). As $\lambda$  increases to $\lambda_2$,  hard samples with larger loss are also taken into consideration as shown in    Fig.  \ref{SPL}.(b). These hard samples could force the model to learn more representative features, and thus properly training these hard samples could improve the generalization ability. However, when $\lambda$ increases to $\lambda_3$,  noisy samples with extremely excessive losses are considered and lead to the degeneration of generalization ability as shown in Fig.  \ref{SPL}.(c).

\subsection{Multi-similarity loss}
 Let $f_{\theta}(\x_i)$ be  an  embedding vector of the sample  $\x_i$ where $f_{\theta}:  \mathbb{R}^M \to \mathbb{R}^m$ means the model with parameters $\theta$. For the training stability,  embedding vectors  have been  normalized. Formally, we define the similarity between two samples as $S_{\x_i,\x_j}= \langle  f_{\theta}(\x_i), f_{\theta}(\x_j) \rangle $, where $\langle   \cdot  ,  \cdot  \rangle $ denotes dot product.  A high similarity between two samples implies that these two samples are close to each other in the embedding space. Besides, in DML, a positive pair means a pair of samples from the same class and a negative pair represents  a pair of samples from different classes.

Similar to contrastive loss  \citep{bromley1993signature} and triplet loss \citep{hermans2017defense}, Multi-Similarity (MS) loss  \citep{MS} is one pair-based DML loss, which pays attention to the  distance information between pairs of samples. In particular,  MS loss utilizes the LogSumExp function to consider the distance information between as many pairs as possible and weights these pairs according to the  General Pair Weighting (GPW) framework \citep{MS}. 
 
Specifically, the MS objective is formulated as:
\begin{equation} \small
\begin{aligned}
	L_{MS}(\theta)=\sum_{c=1}^{C}\frac{1}{N^c} \sum_{a=1}^{N^c} \big( \xi_{\theta}^+(\x_a^c) +\xi_{\theta}^-(\x_a^c) \big),
\end{aligned}
\end{equation}
where $
	\xi_{\theta}^+(\x_a^c) = \frac{1}{\alpha} \log \bigg [1+ \sum\limits_{ \substack{p\in [N^c] \\  p \neq a}} e^{-\alpha(S_{\x_a^c,\x_p^c}-\rho)} \bigg]$,
	$\xi_{\theta}^-(\x_a^c)= \frac{1}{\beta} \log \bigg[1+ \sum\limits_{\substack{k\in [C] \\ k \neq c} }  \sum_{n=1}^{N^k}  e^{\beta(S_{\x_a^c,\x_n^k}-\rho)}\bigg]$.
%\begin{equation}  \label{EqLossInde} \small
%\begin{aligned}
%	&\xi_{\theta}^+(\x_a^c)= \frac{1}{\alpha} \log \bigg [1+ \sum\limits_{ \substack{p\in [N^c] \\  p \neq a}} e^{-\alpha(S_{\x_a^c,\x_p^c}-\rho)} \bigg], \\
%	&\xi_{\theta}^-(\x_a^c)= \frac{1}{\beta} \log \bigg[1+ \sum\limits_{\substack{k\in [C] \\ k \neq c} }  \sum_{n=1}^{N^k}  e^{\beta(S_{\x_a^c,\x_n^k}-\rho)}\bigg].
%\end{aligned}
%\end{equation}
Here, $\alpha, \beta$ and $ \rho$ are fixed hyper-parameters, $\x_a^c$ means the $a$-th sample in  class $c \in [C]$ ,  $N^c$ is  the number of samples in class $c$ and $N=\sum_{c=1}^CN^c$. Moreover, an informative pair mining method is also proposed.  For an anchor sample $\x_a^c$, if a  negative pair $\{\x_a^c, \x_n^k \}, k \in [C], k \neq c, n \in [N^k]$  satisfies the following condition:
\begin{align} \label{ming1}
	S_{\x_a^c, \x_n^k } > \min\limits_{ \substack{ p \in [N^c] \\ p \neq a } } S_{\x_a^c,\x_p^c} - \epsilon,
\end{align}
where $\epsilon$ is a given margin, the negative pair can be considered as an informative negative pair. And an informative positive pair  $\{\x_a^c, \x_p^c \}, p \neq a$ should satisfy:
\begin{align} \label{ming2}
	S_{\x_a^c, \x_p^c} < \max\limits_{ \substack{k \in [C], k \neq c \\ n \in [N^k]}} S_{\x_a^c, \x_n^k} + \epsilon.
\end{align}

\section{Proposed algorithm}
In this section, we provide our balanced self-paced metric learning algorithm followed by its objective function.

\subsection{Objective function} 
As shown in Eq. (1), the classic SPL formulation mainly consists of two parts: the weighted sample loss and the self-paced regularization term. To handle the noisy data, we propose a new denoising multi-similarity formulation:
\begin{equation} \footnotesize  \label{BSPMS}
\begin{aligned} 
	&\mathcal{L}(\theta,\mathbf{w};\lambda)  \\
	  =& \sum_{c=1}^{C} \sum_{a=1}^{N^c} \frac{w_a^c}{N^c}  \bigg (  \frac{\zeta(c)-w_a^c}{N^c-1}  \xi_{\theta}^+(\x_a^c) +   \frac{  \sum\limits_{ \substack{ k \in [C], k \neq c}} }{C-1} \frac{\zeta(k)}{N^k}  \xi_{\theta}^-(\x_a^c) \bigg )  \\
	& - \lambda \sum_{c=1}^{C} \frac{\zeta(c)}{N^c}
	+  \frac{\mu}{C-1}\sum_{c=1}^{C}  \sum_{k=c+1}^{C} \left( \frac{\zeta(c)}{N^c}-\frac{\zeta(k)}{N^k} \right)^2  \\
	  & s.t. \ \  w_a^c \in [0,1],  \  \forall c \in [C], \  a\in [N^c]  
\end{aligned}
\end{equation}
where   $\lambda$ is the age parameter that controls the learning pace in SPL and $\zeta(c)=\sum_{i=1}^{N^c} w^c_i$ means the sum of sample weights in class $c$. Importantly, the last term in Eq. (\ref{BSPMS}) is our new balance regularization term. Specifically, DML problems usually focus on a large number of classes. Thus, we have to face the challenge of the unbalanced average sample weights among classes, \emph{e.g.}, all samples in some classes are assigned the weights of value $1$, while all samples in other classes are assigned the weights of value $0$. Obviously, this situation would reduce the model generalization ability. Fortunately, our balance regularization term can ensure the rationality of the average sample weight by punishing the absolute difference between the average sample weights of different classes.

Compared with the original MS objective, our new formulation can dynamically adjust sample weights through the model feedback to weaken the influence of noisy samples. Moreover, compared with other SPL objectives, we especially consider the balance of the average sample weights under the multi-category condition to ensure stability and reliability of the DML training.

\begin{algorithm} [!htbp] 
	\caption{Balanced self-paced  metric learning}
	\begin{algorithmic}[1]   \label{BSPML}
		\REQUIRE  Maximum iteration $T$,  initial age parameter $ \lambda^0$,  multiplier $c$ and maximum age parameter $\lambda_{\infty}$.\\ 
		\ENSURE The optimal $\theta$.\\ 
		\STATE Initialize $\mathbf{w}^0= \mathbf{1}_N$.
		\FOR { $t=1, \cdots ,T$}
		\STATE Fix $\mathbf{w}$ to optimize $\theta$ through Algorithm \ref{SGDMS}.
		\STATE Fix $\theta$ to optimize $\mathbf{w}$ through Algorithm \ref{DSCD}.
		\STATE $\lambda^{t}=\min (c\lambda^{t-1},\lambda_{\infty})$.
		\ENDFOR
	\end{algorithmic}
\end{algorithm}

\subsection{Balanced self-paced metric learning}
Following the traditional SPL practice,  our  BSPML utilizes the alternative optimization strategy based on two key sub-problems \emph{w.r.t.}   $\mathbf{\theta}$ and $\mathbf{w}$ as shown in Algorithm \ref{BSPML}, where $c$ is the multiplier of  $\lambda$ and $\lambda_{\infty}$ is the maximum value of $\lambda$. 
In the following, we present Algorithms  \ref{SGDMS} and \ref{DSCD} used for solving the two key sub-problems \emph{w.r.t.}   $\mathbf{\theta}$ and $\mathbf{w}$ respectively.

\subsubsection{Algorithm \ref{SGDMS}: 
Mini-batch gradient algorithm for weighted multi-similarity loss}
When we fix $\mathbf{w}$ to optimize $\theta$, our objective function degenerates into a weighted multi-similarity loss formulation that can be optimized by common practice in DML. 
\\ \underline{\textit{Stochastic sampling:}}
We first stochastically sample $P$ classes and then stochastically sample $K$ samples per class.
\\ \underline{\textit{Informative pair mining:}}
We select informative pairs from these $PK$ samples according to Eqs. (\ref{ming1}) and  (\ref{ming2}). In fact, when we use the original MS loss formulation, we inevitably utilize noisy pairs that would produce  noisy gradients as  Fig.  \ref{withoutSPL}.(a) shows. However, in our BSPML algorithm, we exclude noisy samples by reducing these weights under the guidance of the loss value and thus get a satisfactory embedding space, where there exists enough distance between   samples from different classes as shown in Fig. \ref{withoutSPL}.(d).
\\ \underline{\textit{Weighted informative batch loss:}}
For an anchor sample $\x_i$, we define $\hat{P}_{\x_i}$ as the index set of informative positive pairs and  $\hat{N}_{\x_i}$ as the index set of informative negative pairs. Then, the weighted informative batch loss is formed as:
\begin{equation} \footnotesize \label{WBL}
\begin{aligned} 
	& L_{\mathrm{WBL}}(\theta) \\
 = &\frac{1}{PK} \sum_{i=1}^{PK} w_i \bigg \{   \frac{\sum\limits_{p\in \hat{P}_{\x_i} }w_p } {|\hat{P}_{\x_i}| \alpha } \log \bigg[1+\sum_{p\in \hat{P}_{\x_i}} e^{-\alpha(S_{\x_i,\x_p}-\rho)} \bigg]   \\
	&    +  \frac{\sum\limits_{n\in \hat{N}_{\x_i} } w_n }{|\hat{N}_{\x_i}| \beta }  \log \bigg[1+\sum_{n\in \hat{N}_{\x_i} } e^{\beta(S_{\x_i,\x_n}-\rho)} \bigg] \bigg \}.
\end{aligned}
\end{equation}
%\\ \underline{\textit{Algorithm \ref{SGDMS}:}} 

Finally, we summarize the above procedures in  Algorithm \ref{SGDMS}, where the gradient-based method means the optimization method utilizing gradient information such as stochastic gradient descent (SGD) \citep{ketkar2017stochastic} and adaptive moment estimation (Adam) \citep{kingma2017adam}.

\begin{algorithm} [!htbp]   
 	\caption{Mini-batch gradient algorithm for weighted multi-similarity loss} 
 	\begin{algorithmic}[1]    \label{SGDMS}
 		\REQUIRE Maximum iteration $T$, initial model parameter matrix $\theta^0$, class size $P$ and sample size $K$.
 		\ENSURE The optimal  $\theta$.
 		\FOR { $t=1, \cdots ,T$}
 		\STATE Sample $P$ classes stochastically and then sample $K$ samples per class stochastically.
 		\STATE Select informative pair index sets $\hat{P}$ and $\hat{N}$ according to Eqs. (\ref{ming1}) and (\ref{ming2}).
 		\STATE Update $\theta$  using gradient-based method with the loss $L_{\mathrm{WBL}}$, \emph{i.e.}, Eq. (\ref{WBL}).
 		\ENDFOR
 	\end{algorithmic}
 \end{algorithm}
 
 \begin{algorithm} [!htbp]   
 	\caption{Doubly stochastic projection coordinate gradient method } \label{DSCD}
 	\begin{algorithmic}[1]   
 		\REQUIRE Maximum iteration $T$, initial sample weight vector $\mathbf{w}^0$, learning rate $\gamma$,   class size $P$ and weight size $K$.
 		\ENSURE The optimal  $\mathbf{w}$.
 		\FOR{$t=1,\cdots,T$}
 		\STATE Sample one weight $w_a^c,c \in [C],a \in [N^c]$ stochastically.  
 		\STATE Sample $K$ weights different from $w_a^c$ in class $c$ stochastically, sample $P$ classes different from class $c$ and $K$ weights per class stochastically.
 		\STATE Calculate the stochastic gradient $G(w_a^c)$.
 		\STATE Update $\mathbf{w}^t =\mathcal{P}_{[0,1]^N}(\mathbf{w}^{t-1} - \gamma^t G(w_a^c) \mathbf{e}_a^c)$.
 		\ENDFOR
 	\end{algorithmic}
 \end{algorithm}
 
\subsubsection{Algorithm \ref{DSCD}: Doubly stochastic projection coordinate gradient method}
When we fix  $\theta$ to optimize $\mathbf{w}$,   the sub-problem  \emph{w.r.t.} $\w$  is  a more complex quadratic  problem  compared with the one in  existing SPL problems, \emph{e.g.}, Eq. (\ref{EqclassicSPL}). To solve this  complex  sub-problem  efficiently, we propose a  doubly stochastic projection coordinate gradient method. 
\\ \underline{\textit{Doubly stochastic sampling:}}
Specifically, for an anchor weight $w_a^c$, we first stochastically select $K$  weights different from $w_a^c$ in class $c$. Then, we  stochastically select $P$  classes different from $c$ and  select $K$ weights per class. 
 \\ \underline{\textit{Stochastic gradients}}: Under the circumstance, we show the stochastic gradient $G( w^c_a)$ regarding to a particular coordinate $w^c_a$ as follows:
\begin{equation}\label{UE} \small
\begin{aligned} 
&G(w_a^c)= \frac{1}{N^c} \big( G_p(w_a^c)+G_n(w_a^c)+G_b(w_a^c)   -  \lambda   \big),\\
&G_p(w_a^c)=\frac{1}{K} \sum_{p=1}^K w_p^c \big( \xi_{\theta}^+(\x_p^c) +     \xi_{\theta}^+(\x_a^c)  \big),     \\ 
&G_n(w_a^c)= \frac{1}{P} \sum_{k=1}^P   \frac{1}{K}  \sum_{n=1}^{K}   w_n^k   \big(  \xi_{\theta}^- (\x_n^k)  + \xi_{\theta}^- (\x_a^c)   \big),  \\
&G_b( w_a^c)= 2 \mu  \left ( \frac{\sum_{i=1}^{N^c} w^c_i}{N^c} - \frac{1}{C-1} \sum_{\substack{ k \in [C] \\ k \neq c }} \frac{\sum_{i=1}^{N^k} w^k_i}{N^k}  \right).
\end{aligned}\end{equation}
%Because $\theta$ is fixed in Eq. (\ref{UE}), we have that both $\xi_{\theta}^+$  and $\xi_{\theta}^-$  are constants, and $\mathbb{E}[ G(w_a^c)]  =  \frac{\partial \mathcal{L}(\mathbf{w};\theta,\lambda)}{\partial w_a^c}$.

The gradient $G$ in Eq. (\ref{UE}) is  composed of four terms. Both $G_p$  and $G_n$ are related to the similarity between samples,  and  they imply that if one sample owns the high similarity  to  samples from  different classes and the low similarity to  samples from the same class,  our algorithm would  exclude this sample from the training  by gradually reducing its weight. Moreover, the $G_b$ is generated by our proposed balance regularization term. As we expected, a sample will be assigned a larger weight if the average sample weight $\frac{\sum_{i=1}^{N^c} w^c_i}{N^c}$ of its class is low, or the average sample weight $\frac{1}{C-1} \sum_{\substack{ k \in [C], k \neq c }} \frac{\sum_{i=1}^{N^k} w^k_i}{N^k}$ of all other classes is high. Finally, the last term of Eq. (\ref{UE}) is the age parameter $\lambda$, which controls the learning pace in SPL. With increased $\lambda$, our algorithm tends to assign samples larger weights.

The doubly stochastic projection coordinate gradient method is summarized in Algorithm \ref{DSCD}, where $\gamma$ means the learning rate,  $\mathcal{P}_{S}$ is the projection operation to the set $S$ and $\mathbf{e}_a^c$ is one unit vector where the coordinate with the same index  as $w_a^c$  holds the  value of $1$.

\section{Theoretical analysis}
In this section, we prove the convergence of  Algorithms \ref{DSCD} and \ref{BSPML}. All the proof details are available in Appendix.

\noindent \textbf{Convergence of Algorithm \ref{DSCD}}:  For Algorithm \ref{DSCD}, let  
$w_{j(t)}$ mean the coordinate selected in $t$-iteration and $\mathcal{B}(t)$ mean the mini-batch sampled  in $t$-iteration.  Then, we define
$$\nabla_{j(t)} \mathcal{L}(\w^t;\theta,\lambda) = \frac{\partial \mathcal{L}(\mathbf{w}^t;\theta,\lambda)}{\partial w_{j(t)}}.$$  
Next, we  introduce the necessary assumption and the definition of the projected gradient.

\begin{assumption}  \label{assumptionVar}
For any $t \in \mathbb{N}$, we have
\begin{align} \label{EqassumptionVar}
	\mathbb{E}_{\mathcal{B}(t)} \big[  | G(w_{j(t)}) -  \nabla_{j(t)} \mathcal{L}(\w^t;\theta,\lambda)    |^2  \big] \leq (\sigma^t)^2, 
\end{align}
where  $\sigma^t>0$ is some constant. 
\end{assumption}

\begin{definition}[Projected gradient \citep{ghadimi2016mini}] \label{definitionProGra}
	Let $S$ be a closed convex set with dimension $N$ and the projected gradient  is defined as:
	\begin{align} \label{EqdefinitionProGra}
		\mathcal{G}_S(\w,\mathbf{g},\gamma)=\frac{1}{\gamma}(\w-\mathcal{P_S}(\w-\gamma \mathbf{g} )),
	\end{align}
	where $\w \in S$, $\mathbf{g} \in \mathbb{R}^N$ and  $\gamma \in \mathbb{R}$.
\end{definition}

Assumption \ref{assumptionVar} is a common assumption in stochastic optimization \citep{gu2019scalable}, which bounds the error between the stochastic gradient and the full gradient. Besides, based on Definition \ref{definitionProGra}, we have that
\begin{align}
    \mathcal{G}_{[0,1]^N}(\w^t,G(w_{j(t)}) \e_{j(t)},\gamma^t)=\frac{ \w^{t} - \w^{t+1} }{\gamma^t} := \hat{D}^t
\end{align}
means the projected gradient generated  in $t$-iteration, where $\mathbf{e}_i$ is one unit vector with  $i$-th coordinate holding the value of $1$.
Based on these, we provide the convergence of Algorithm \ref{DSCD} in Theorem \ref{thm1}.
\begin{theorem}  \label{thm1}
When  Assumption  \ref{assumptionVar} holds and we use Algorithm \ref{DSCD} to optimize $\w$, $L_{\mathrm{max}}>0$ is the maximum lipschitz constant of sub-problems \emph{w.r.t.} single coordinate $w_i, i \in [N]$, stepsizes  $\{\gamma^t\}_{t=1}^{\infty}$ satisfy $0< \gamma^{t+1} \le \gamma^{t} < \frac{2}{L_{\mathrm{max}}}$ and 
\begin{equation} \label{stepsize}
	\sum_{t=1}^{\infty} \gamma^t = + \infty, \quad \sum_{t=1}^{\infty} \gamma^t (\sigma^t)^2 < \infty,
\end{equation}
then there exists an index sub-sequence $\mathcal{K}$ such that
\begin{align}
	\lim\limits_{\substack{t \to \infty \\ t\in \mathcal{K} }}\mathbb{E}  ||\hat{D}^t ||_2=0.
\end{align}
%which implies Algorithm \ref{DSCD} approaches to a stationary point of the sub-problem% \emph{w.r.t.} $\w$.
\end{theorem}

\begin{remark}
The above theorem shows that Algorithm \ref{DSCD} approaches to a stationary point of the sub-problem \emph{w.r.t.} sample weights. It indicates that our algorithm can assign the  appropriate sample weight  to guild the sample selection.
\end{remark}

\noindent \textbf{Convergence of Algorithm \ref{BSPML}}:
Before providing the convergence of our BSPML algorithm, \emph{i.e.}, Algorithm \ref{BSPML},  we introduce the following assumption:
\begin{assumption} \label{assumptiongradient}
	If we call the gradient-based algorithm to minimize $F(\mathbf{X})$ with  an initial solution $\mathbf{X}^0$, we have that $F(\mathbf{X}^0) \geq F(\mathbf{X}^t)$   with  a large enough number $t$ of  iterations.
\end{assumption}

It is easy to verify that Assumption \ref{assumptiongradient} is a basic requirement for a  gradient-based optimizer no matter whether the optimization  objective is convex or not.
 At this time, we give the convergence  of our BSPML as follows.

\begin{theorem} \label{Converge}
If Assumption \ref{assumptiongradient} holds,  the objective function sequence $\{ \mathcal{L}(\theta^{t},\mathbf{w}^{t};\lambda^{t}) \}_{t=1}^T$ generated by the BSPML algorithm converges  with the following property: 
\begin{align*}
    \lim\limits_{t \to \infty}  || \mathcal{L}(\theta^{t},\mathbf{w}^{t};\lambda^{t}) - \mathcal{L}(\theta^{t-1},\mathbf{w}^{t-1};\lambda^{t-1}) || =0.
\end{align*}
\end{theorem}
\begin{remark}
The above theorem shows that when the number of iterations is large enough, our objective function can converge to a fixed value by using our BSPML algorithm.
\end{remark}

\begin{table*}[!htbp]   \small
\centering
\setlength{\tabcolsep}{3pt} 
\begin{tabular}{|c|ccccccc|ccccccc|}
\hline
                   & \multicolumn{7}{c|}{CUB}                                                                                                    & \multicolumn{7}{c|}{Cars}                                                                                                    \\ \cline{2-15} 
                   & R@1             & R@2             & R@4             & R@8             & R@16            & R@32            & NMI             & R@1             & R@2             & R@4             & R@8             & R@16            & R@32            & NMI             \\ \hline
Contrastive        & 49.6            & 62.1            & 73.3            & 82.7            & $\textbf{89.8}$ & 94.6            & 57.7            & 56.4            & 68.1            & 77.2            & 85.7            & 91.3            & 95.6            & 51.8            \\
Triplet$_{\text{smooth}}$ & 42.0            & 54.5            & 66.4            & 77.6            & 86.5            & 92.7            & 52.3            & 37.4            & 49.9            & 62.3            & 73.7            & 83.6            & 90.7            & 47.9            \\
Margin             & 38.0            & 49.9            & 62.7            & 74.6            & 84.8            & 91.4            & 51.4            & 42.8            & 54.6            & 66.5            & 77.7            & 86.7            & 93.1            & 49.4            \\
FastAP             & 42.9            & 55.0            & 67.3            & 78.0            & 86.8            & 92.9            & 52.7            & 33.6            & 46.7            & 59.5            & 71.6            & 82.3            & 90.6            & 49.4            \\
MS                 & 49.2            & 61.5            & 73.5            & 83.1            & 89.5            & 94.6            & 58.2            & 58.1            & 70.2            & 80.0            & 87.5            & 92.6            & 96.1            & 54.4            \\
RDML               & 43.5            & 56.8            & 69.1            & 80.1            & 88.5            & 93.9            & 53.1            & 37.7            & 51.5            & 64.7            & 76.1            & 85.5            & 92.3            & 49.1            \\
Proxy-Anchor       & 48.9            & 61.6            & 72.9            & 82.7            & 89.4            & 94.6            & 57.9            & 58.5            & 70.4            & 80.2            & 87.4            & 92.7            & $\textbf{96.5}$ & 54.6            \\
SNR                & 47.8            & 61.5            & 72.9            & 82.6            & 89.3            & 94.3            & 56.4            & 56.0            & 67.9            & 78.1            & 86.1            & 91.9            & 95.5            & 51.2            \\
BSPML              & $\textbf{50.6}$ & $\textbf{62.8}$ & $\textbf{74.1}$ & $\textbf{83.5}$ & 89.7            & $\textbf{94.9}$ & $\textbf{58.9}$ & $\textbf{59.6}$ & $\textbf{71.4}$ & $\textbf{80.3}$ & $\textbf{87.6}$ & $\textbf{92.9}$ & 96.4            & $\textbf{55.4}$ \\ \hline
                   & \multicolumn{7}{c|}{In-shop}                                                                                                & \multicolumn{7}{c|}{SOP}                                                                                                     \\ \cline{2-15} 
                   & R@1             & R@2             & R@4             & R@8             & R@16            & R@32            & NMI             & R@1             & R@8             & R@16            & R@32            & R@64            & R@128           & NMI             \\ \hline
Contrastive        & 73.1            & 80.9            & 86.8            & 90.8            & 93.4            & 95.5            & 82.9            & 65.0            & 79.5            & 83.1            & 86.2            & 88.7            & 91.1            & 88.3            \\
Triplet$_{\text{smooth}}$ & 67.6            & 77.0            & 84.7            & 89.8            & 93.5            & 96.0            & 83.2            & 61.6            & 77.4            & 81.6            & 85.3            & 88.4            & 91.1            & 87.8            \\
Margin             & 55.7            & 66.5            & 75.9            & 83.2            & 88.9            & 93.0            & 80.1            & 51.4            & 69.9            & 75.3            & 80.4            & 84.8            & 88.7            & 86.2            \\
FastAP             & 64.6            & 74.2            & 82.3            & 88.1            & 92.1            & 94.7            & 82.3            & 58.6            & 76.0            & 80.5            & 84.6            & 88.0            & 91.0            & 87.5            \\
MS                 & 73.4            & 80.7            & 86.4            & 90.6            & 93.4            & 95.3            & 84.7            & 65.8            & 79.4            & 83.0            & 86.2            & 89.2            & 91.7            & 88.6            \\
RDML               & 69.6            & 78.7            & 85.4            & 90.3            & 93.5            & 95.7            & 83.0            & 62.8            & 78.0            & 82.1            & 85.6            & 88.7            & 90.7            & 88.1            \\
Proxy-Anchor       & 74.2            & 81.2            & 86.9            & 91.0            & 93.7            & 95.5            & 85.0            & $\textbf{66.3}$ & 80.2            & 83.3            & 86.2            & 88.7            & 90.9            & 88.7            \\
SNR                & 72.7            & 80.6            & 87.1            & 91.3            & 94.1            & $\textbf{96.2}$ & 83.6            & 65.3            & 79.0            & 82.3            & 85.5            & 88.3            & 90.6            & 88.3            \\
BSPML              & $\textbf{75.0}$ & $\textbf{82.1}$ & $\textbf{87.8}$ & $\textbf{91.6}$ & $\textbf{94.2}$ & 96.0            & $\textbf{85.9}$ & $\textbf{66.3}$ & $\textbf{80.9}$ & $\textbf{84.4}$ & $\textbf{87.5}$ & $\textbf{90.1}$ & $\textbf{92.5}$ & $\textbf{89.1}$ \\ \hline
\end{tabular}
\vspace{-4pt}
 \caption{ Retrieval and clustering performance  (\%) on original data sets. } \label{clean} 
 \vspace{-8pt}
\end{table*}

\begin{table*}[!h]  \footnotesize
\centering
\setlength{\tabcolsep}{2pt} 
\begin{tabular}{|c|c|ccccccccc|}
\hline
                         & NR & Contrastive      & Triplet$_{\text{smooth}}$ & Margin           & FastAP           & MS               & RDML             & Proxy-Anchor     & SNR              & BSPML                     \\ \hline
\multirow{3}{*}{CUB}     & 10\%       & 48.26$\pm$0.21 & 39.69$\pm$0.17   & 37.72$\pm$0.18 & 41.57$\pm$0.13 & 48.46$\pm$0.16 & 41.59$\pm$0.15 & 48.31$\pm$0.12 & 45.95$\pm$0.19 & $\mathbf{50.08\pm0.18}$ \\
                         & 20\%       & 47.79$\pm$0.19 & 36.25$\pm$0.17   & 35.06$\pm$0.20 & 38.80$\pm$0.15 & 47.68$\pm$0.16 & 39.63$\pm$0.16 & 47.62$\pm$0.14 & 45.16$\pm$0.17 & $\mathbf{49.38\pm0.20}$ \\
                         & 30\%       & 46.25$\pm$0.24 & 33.92$\pm$0.20   & 34.36$\pm$0.18 & 37.10$\pm$0.15 & 46.25$\pm$0.14 & 37.57$\pm$0.16 & 46.66$\pm$0.15 & 44.43$\pm$0.21 & $\mathbf{48.74\pm0.19}$ \\ \hline
\multirow{3}{*}{Car}     & 10\%       & 55.24$\pm$0.26 & 36.58$\pm$0.18   & 40.24$\pm$0.18 & 32.33$\pm$0.12 & 57.11$\pm$0.15 & 37.09$\pm$0.16 & 57.84$\pm$0.15 & 55.04$\pm$0.17 & $\mathbf{58.78\pm0.18}$ \\
                         & 20\%       & 54.55$\pm$0.24 & 35.76$\pm$0.17   & 38.49$\pm$0.21 & 31.68$\pm$0.10 & 56.37$\pm$0.17 & 36.32$\pm$0.15 & 57.18$\pm$0.15 & 53.49$\pm$0.18 & $\mathbf{58.24\pm0.16}$ \\
                         & 30\%       & 52.94$\pm$0.22 & 34.81$\pm$0.19   & 37.67$\pm$0.17 & 31.04$\pm$0.10 & 55.14$\pm$0.18 & 35.55$\pm$0.15 & 56.06$\pm$0.16 & 51.51$\pm$0.20 & $\mathbf{57.38\pm0.19}$ \\ \hline
\multirow{3}{*}{In-shop} & 10\%       & 69.59$\pm$0.29 & 59.69$\pm$0.21   & 50.33$\pm$0.20 & 56.28$\pm$0.18 & 70.33$\pm$0.25 & 62.50$\pm$0.22 & 70.51$\pm$0.17 & 67.24$\pm$0.25 & $\mathbf{71.44\pm0.23}$ \\
                         & 20\%       & 63.34$\pm$0.27 & 53.20$\pm$0.18   & 45.74$\pm$0.21 & 52.03$\pm$0.13 & 65.68$\pm$0.20 & 57.73$\pm$0.20 & 66.38$\pm$0.17 & 60.36$\pm$0.23 & $\mathbf{67.91\pm0.25}$ \\
                         & 30\%       & 56.76$\pm$0.22 & 48.35$\pm$0.17   & 43.16$\pm$0.15 & 49.37$\pm$0.13 & 58.85$\pm$0.24 & 53.97$\pm$0.21 & 60.25$\pm$0.15 & 53.62$\pm$0.23 & $\mathbf{62.04\pm0.23}$ \\ \hline
\end{tabular}
\vspace{-4pt}
\caption{ Recall@$1$ results (\%) with the corresponding standard deviation on noisy data sets with different Noise Ratios (NR). } \label{noise}
\vspace{-8pt}
\end{table*}

\section{Experiments}
In this section, we present experimental results  to demonstrate the superiority of our BSPML algorithm.

\subsection{Experimental setup}
\noindent\textbf{Data sets:}  We  conduct the experiments  on  four standard  data sets: CUB200 \citep{CUB},  Cars-196 \citep{Car}, In-Shop \citep{Inshop} and   Stanford  Online  Products (SOP) \citep{SOP}.  For all data sets, we use half of the classes for training and the rest for testing. In addition, to test the robustness of all methods, we construct artificial noisy data sets. Specifically, we stochastically select samples from each class in the training set and change their labels to other classes. Moreover, we try our best to ensure that noisy samples exist in each class,  and we conduct experiments with different noise ratios (from 10\% to 30\%). Since there are only a few samples in most classes of the SOP data set, making it noisy would seriously break the nature of this data set. Thus,  we did not conduct experiments on the noisy SOP data set. \\
\noindent\textbf{Compared algorithms:}  We compare our algorithm with  classic DML algorithms, \emph{i.e.}, contrastive loss \citep{hadsell2006dimensionality} and triplet$_{\text{smooth}}$ loss \citep{hermans2017defense},  state-of-the-art DML algorithms, \emph{i.e.}, margin loss \citep{wu2017sampling}, FastAP loss \citep{cakir2019deep}  and MS loss \citep{MS}, and robust DML algorithms, \emph{i.e.}, RDML loss \citep{rescaled}, Proxy-Anchor  loss \citep{proxyanchor} and  SNR loss \citep{yuan2019signal}. \\
\noindent\textbf{Design of experiments:} All experiments are conducted at least 10 times on a PC with 48 2.3GHz cores, 80GB RAM and 4 Nvidia 1080ti GPUs.  All algorithms are implemented based on the open PyTorch package \citep{musgrave2020metric} using the same network structure with the embedding size $512$. All the network parameters are optimized by  SGD with the learning rate $5e\text{-}6$ and the batch size is set to $64$. Specifically,  one batch is constructed by first sampling $P=16$ classes, and then sampling $K=4$ images for each class. For contrastive loss and SNR loss, the positive margin is set to $1$ and the negative margin is set to $0$. 
As original paper shows, we set  $\alpha=0.2, \beta^{(0)}=1.2$ and $\beta^{(class)}=\beta^{(img)}=0$ for margin loss. For FastAP loss, the number of soft histogram bins is set to 10 recommended by the authors.  $\eta$ is selected from $\{0.5, 1, 2, 3, 4\}$ for RDML loss. For Proxy-Anchor loss, $\alpha=32, \delta=0.1$ and all proxies are initialized using a normal distribution. For MS loss and the MS loss part of our BSPML, $\epsilon$ is set to 0.1 and $\alpha=2,\rho=1, \beta=50$. And  for the balanced self-paced part of our BSPML,  $\lambda_{\infty}$ is tuned  in [1,5] and $\mu=\lambda_{\infty}$. 

Considering hyper-parameters $\mu$ and $\lambda$, we design experiments to analyze their roles. Note that we introduce the Mean of Average Weights of classes (MAW) and the Standard Deviation of Average Weights of classes (SDAW):
\begin{equation} \label{EqMAWandEqSDAW} \small
\begin{aligned}
    & \text{MAW}=\frac{1}{C}   \sum_{c=1}^C \frac{\sum_{i=1}^{N^c} w_i^c}{N^c},\\
    & \text{SDAW}=  \bigg( \frac{1}{C} \sum_{c=1}^C \big(\frac{\sum_{i=1}^{N^c} w_i^c}{N^c} - \text{MAW}\big)^2 \bigg)^{\frac{1}{2}}. 
\end{aligned}
\end{equation}
 MAW  represents the average weight of all samples and SDAW  implies the balance degree between the average sample weights of classes. The smaller the value of SDAW, the more balanced the average sample weights of classes. Moreover, to verify the feasibility of each part of our BSPML, we carry out ablation experiments with varying embedding sizes $\{64, 128, 256, 512\}$ on noisy data sets.  

For the retrieval task, all algorithms are evaluated by the standard performance metric Recall@$K$. To calculate Recall@$K$,  each testing sample first retrieves $K$ nearest neighbors from the test set and receives a score $1$ if a sample of the same class is retrieved among the $K$ nearest neighbors. Considering the clustering performance, we utilize the normalized mutual information (NMI) score:
$\text{NMI}( \Omega, \mathcal{C}) = \frac{2  I( \Omega,\mathcal{C})}{H( \Omega)+H(\mathcal{C})} ,$
where $ \Omega$ denotes the real clustering result and $\mathcal{C}$ denotes the set of clusters obtained by K-means. Here,  $I(\cdot)$ represents the mutual information and $H(\cdot)$ represents the entropy.  

\begin{figure*}[!htp] \small
	\centering
	\subfigure[CUB]{
		\centering
		\includegraphics[width=2.0in]{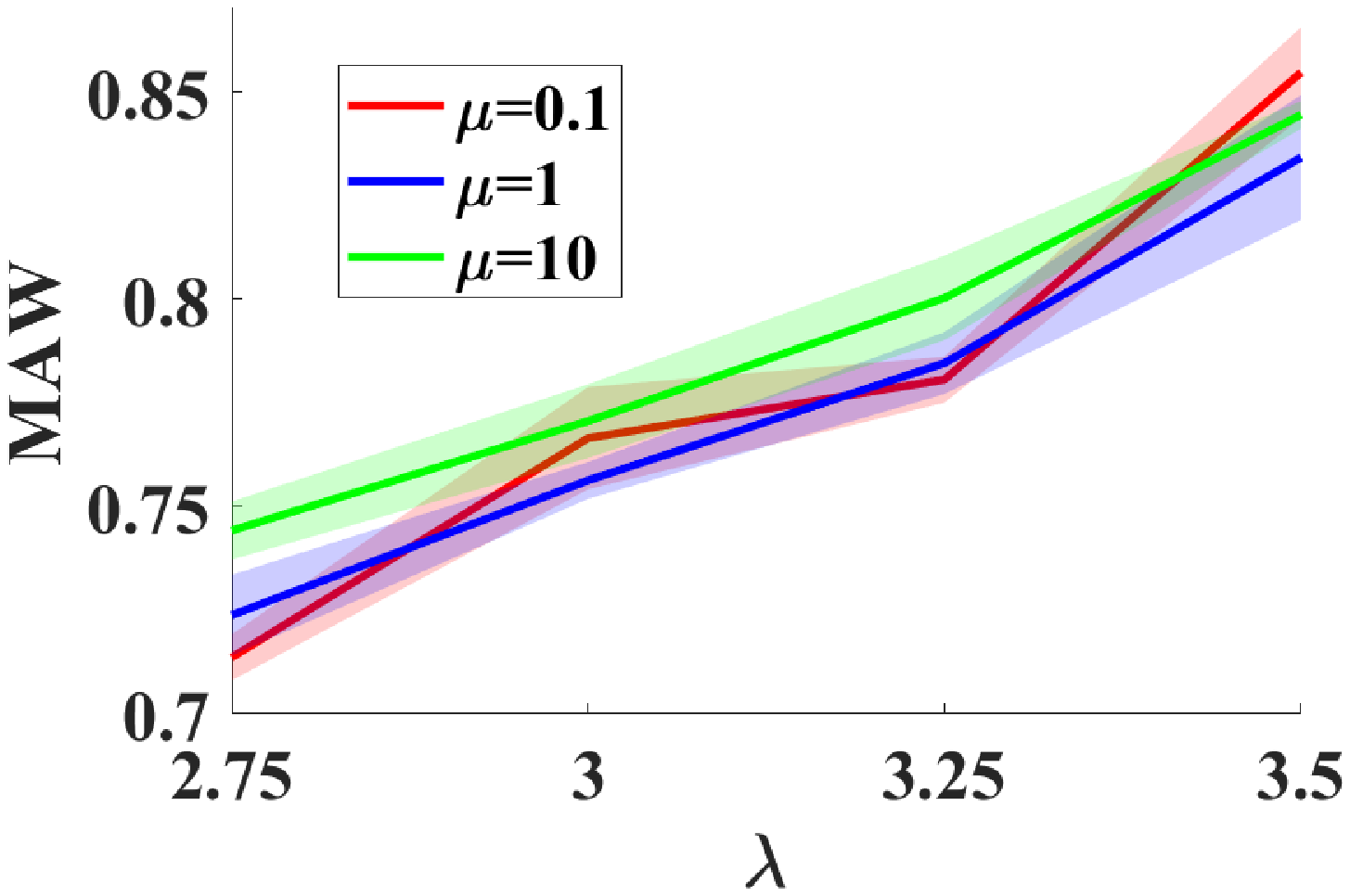}
	} 
	\subfigure[Car]{
		\centering
		\includegraphics[width=2.0in]{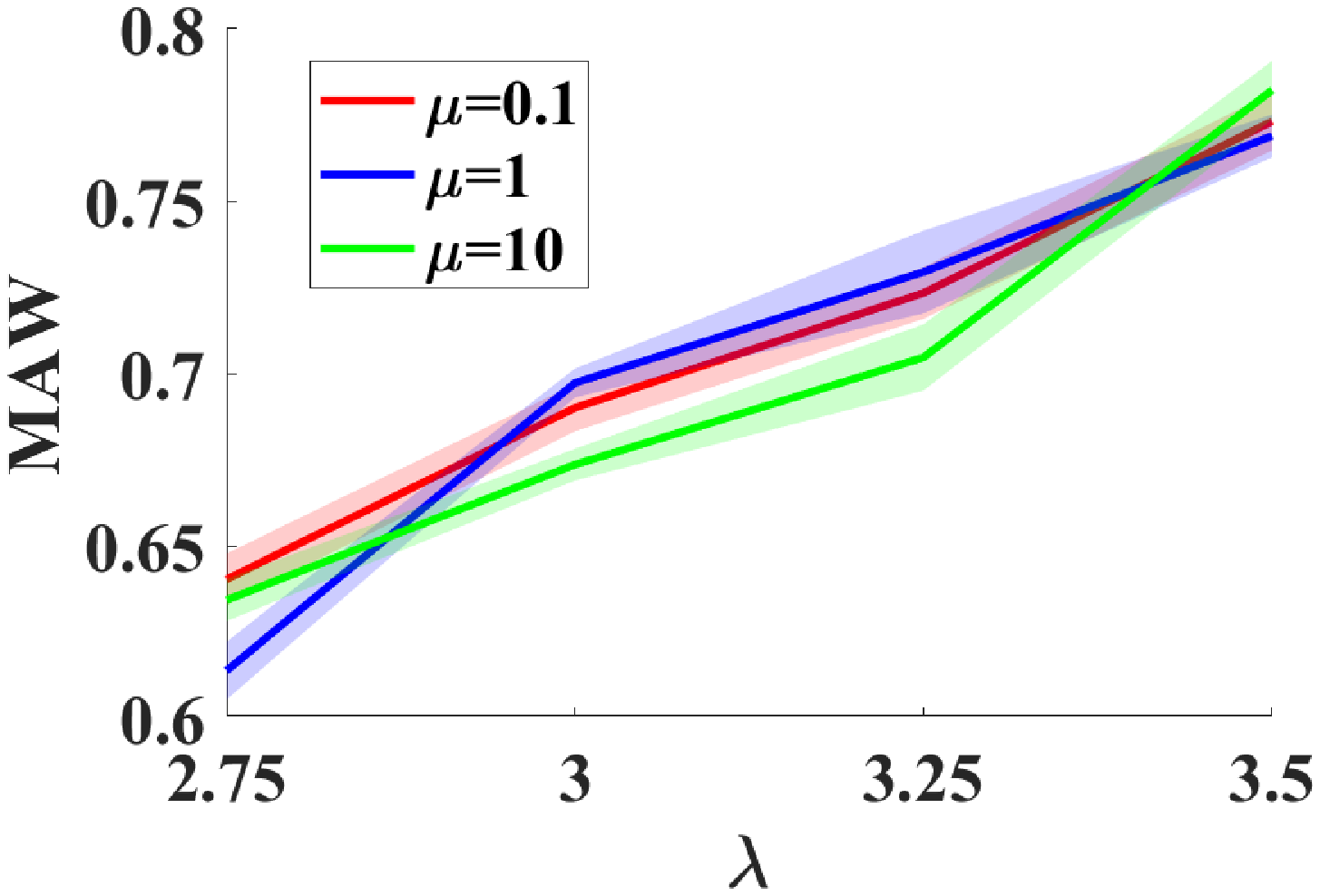}
	}
	\subfigure[In-shop]{
		\centering
		\includegraphics[width=2.0in]{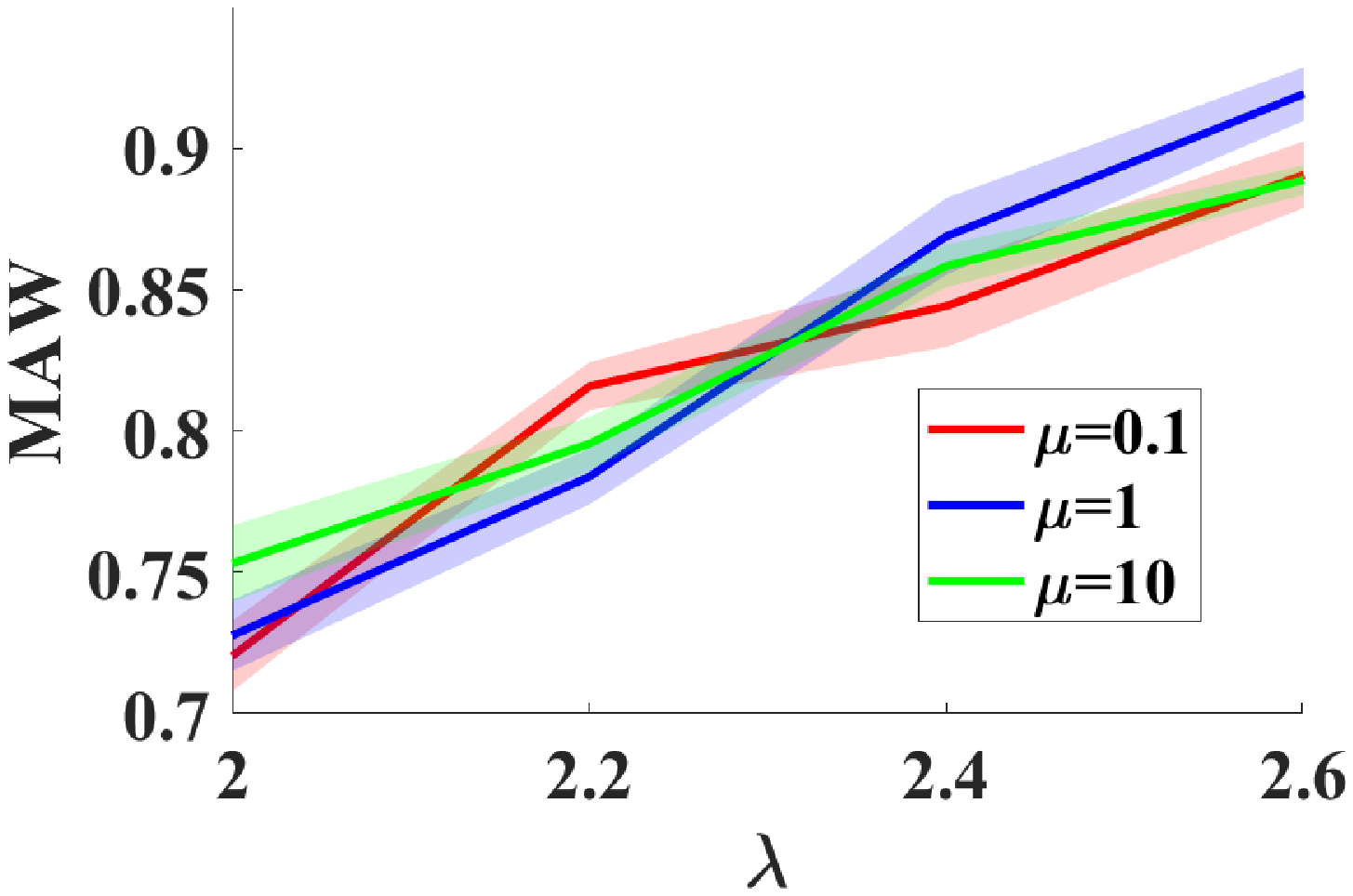}
	}
	\vspace{-4pt}
	\caption{MAW  results with different values of $\mu$ and $\lambda$  on  data sets with 20\% noisy samples.} \label{parameterLambda}
		\vspace{-8pt}
\end{figure*}

\begin{figure*}[!h]  \small
	\centering
	\subfigure[CUB]{
		\centering
		\includegraphics[width=2.0in]{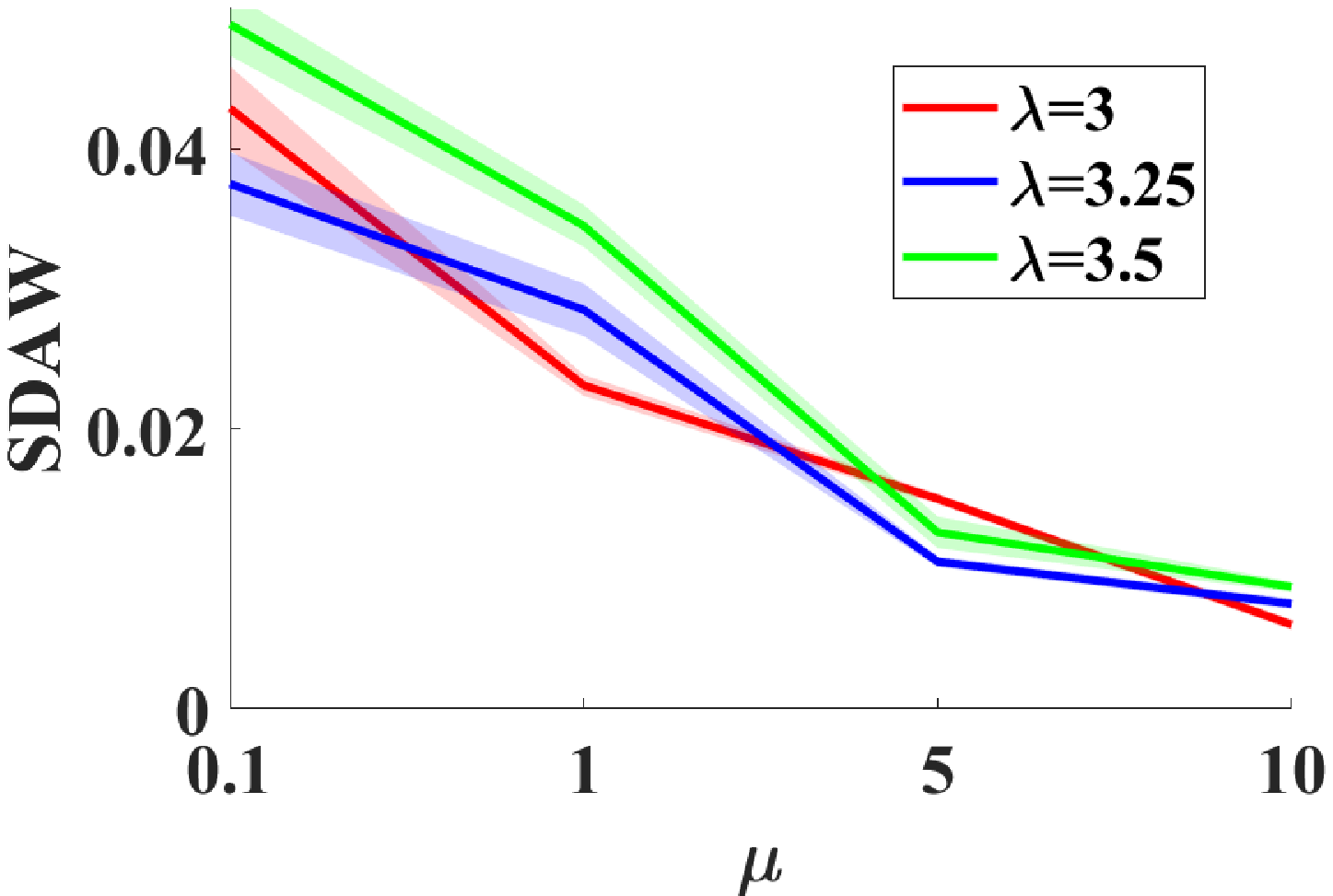}
	}
	\subfigure[Car]{
		\centering
		\includegraphics[width=2.0in]{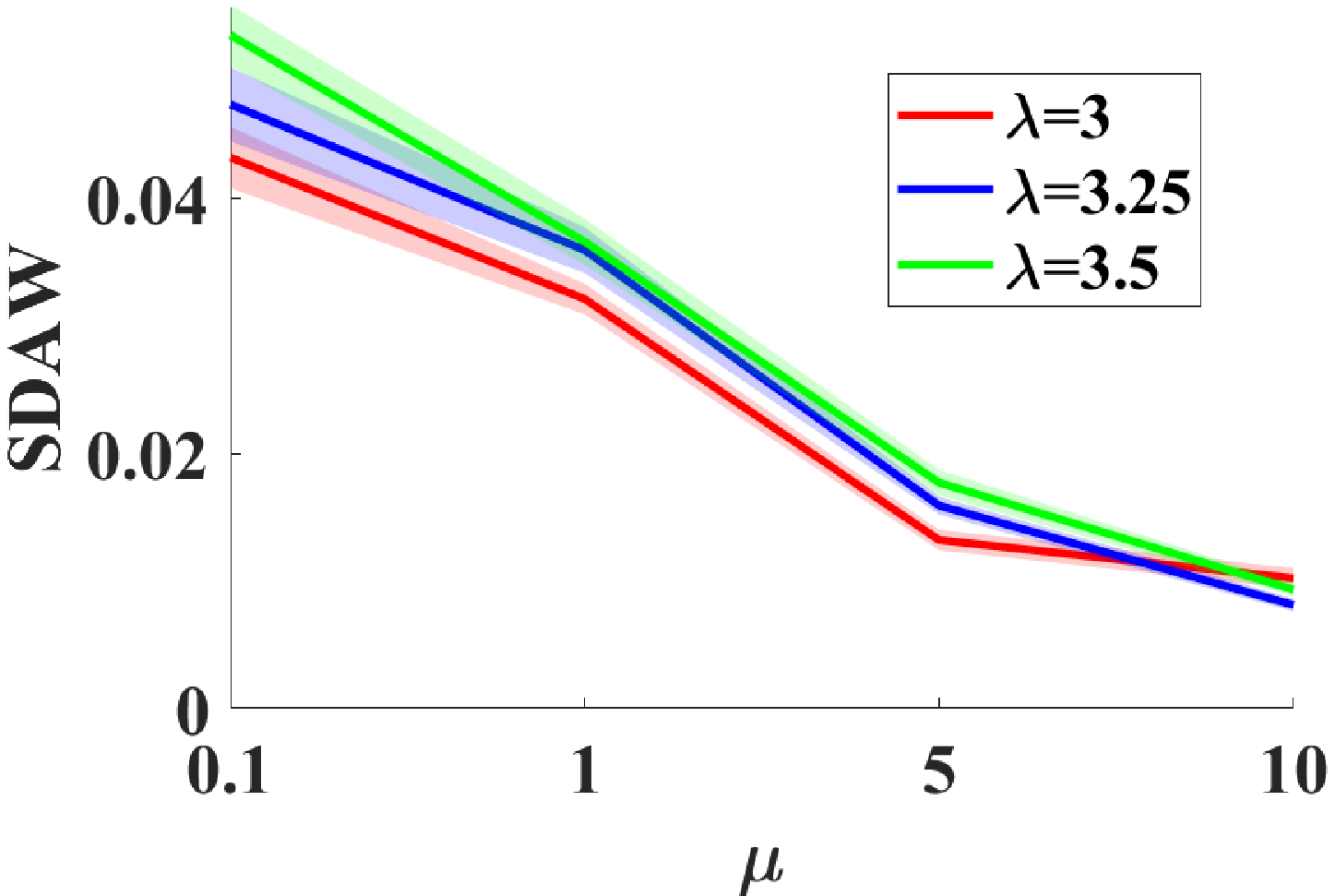}
	}
	\subfigure[In-shop]{
		\centering
		\includegraphics[width=2.0in]{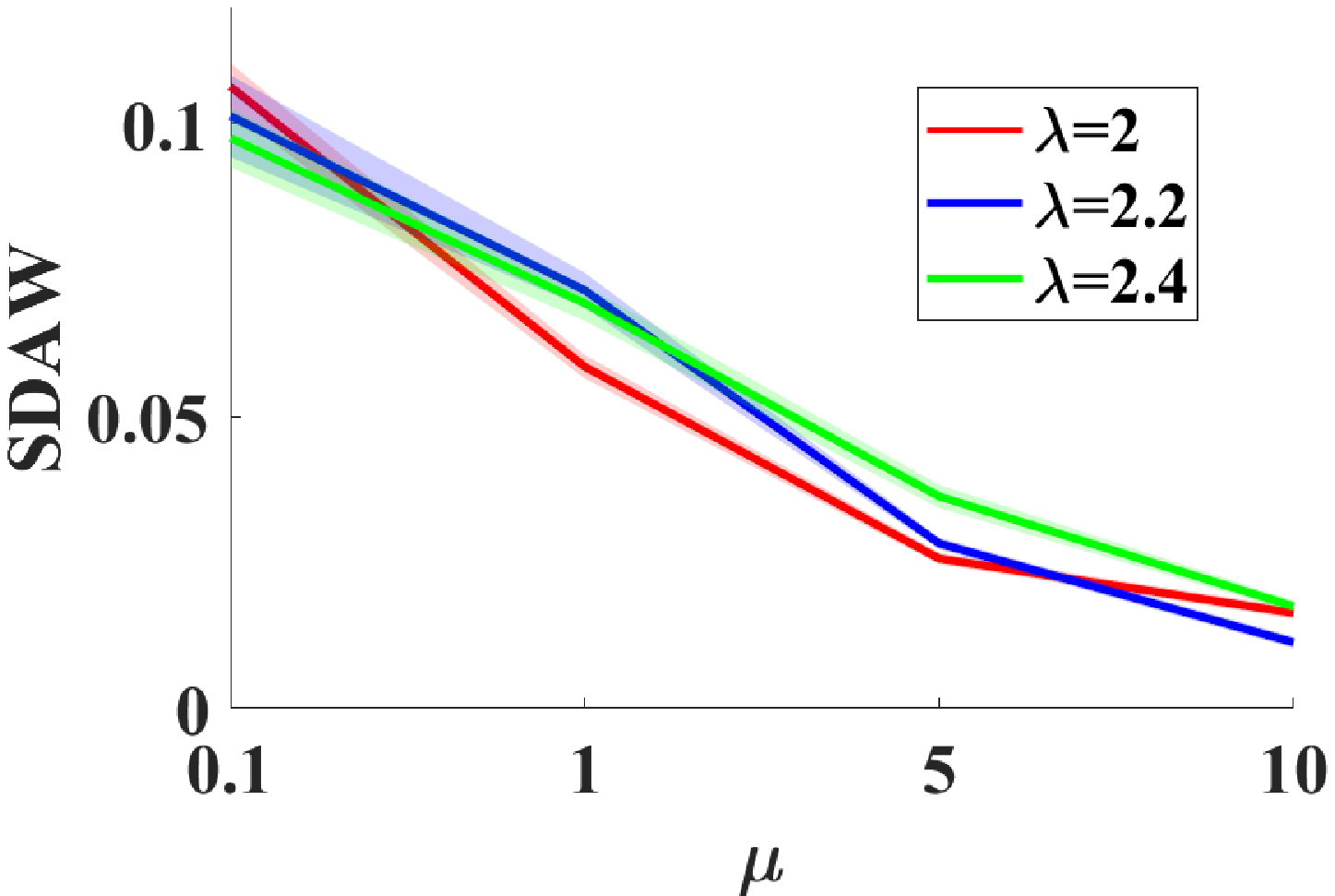}
	}
		\vspace{-4pt}
	\caption{SDAW  results with different values of $\mu$ and $\lambda$  on  data sets with 20\% noisy samples.} \label{parameterMu}
		\vspace{-8pt}
\end{figure*}
\begin{figure*}[!h]  \small
	\centering
	\subfigure[CUB]{
		\centering
		\includegraphics[width=2.0in]{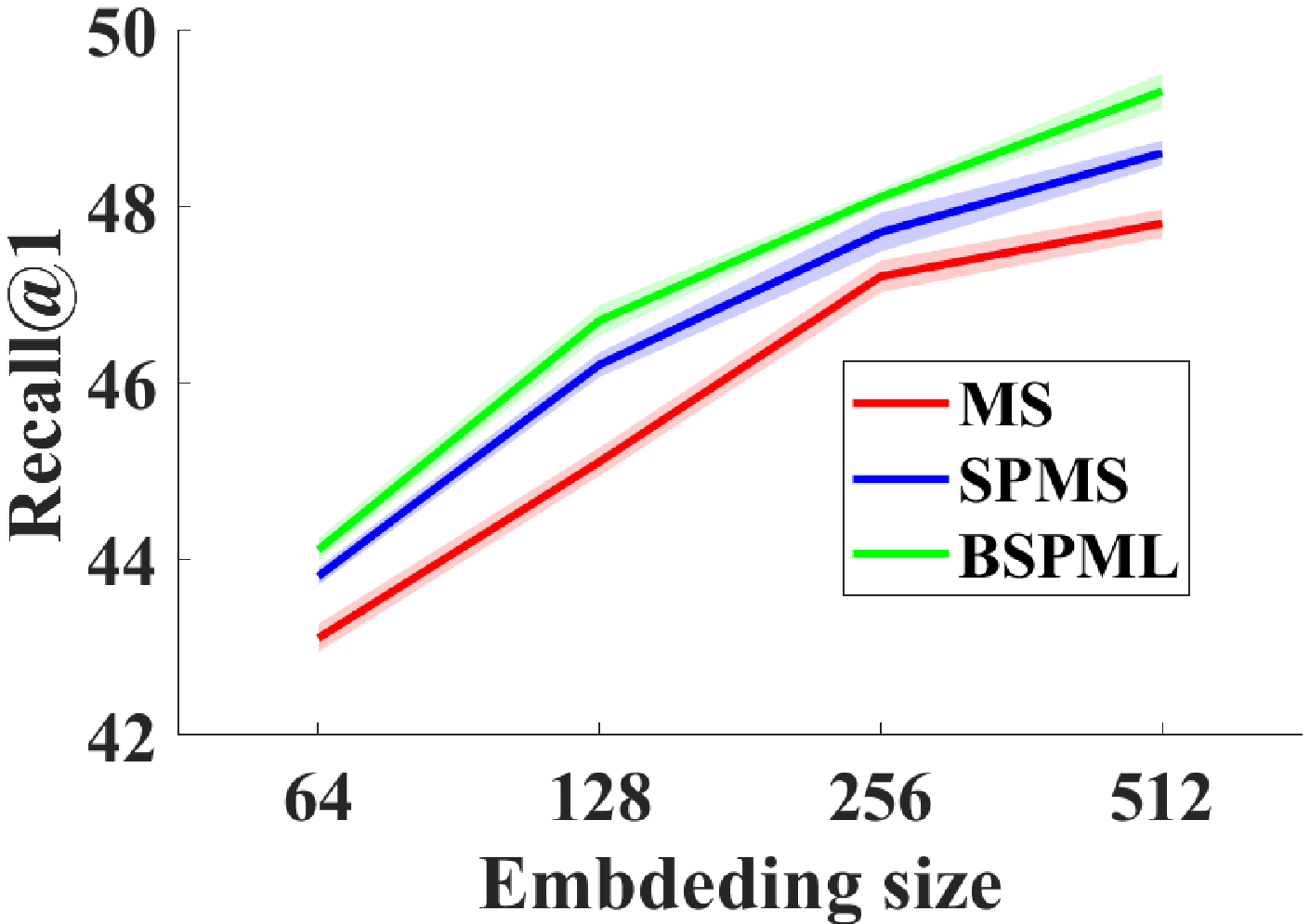}
	}
	\subfigure[Car]{
		\centering
		\includegraphics[width=2.0in]{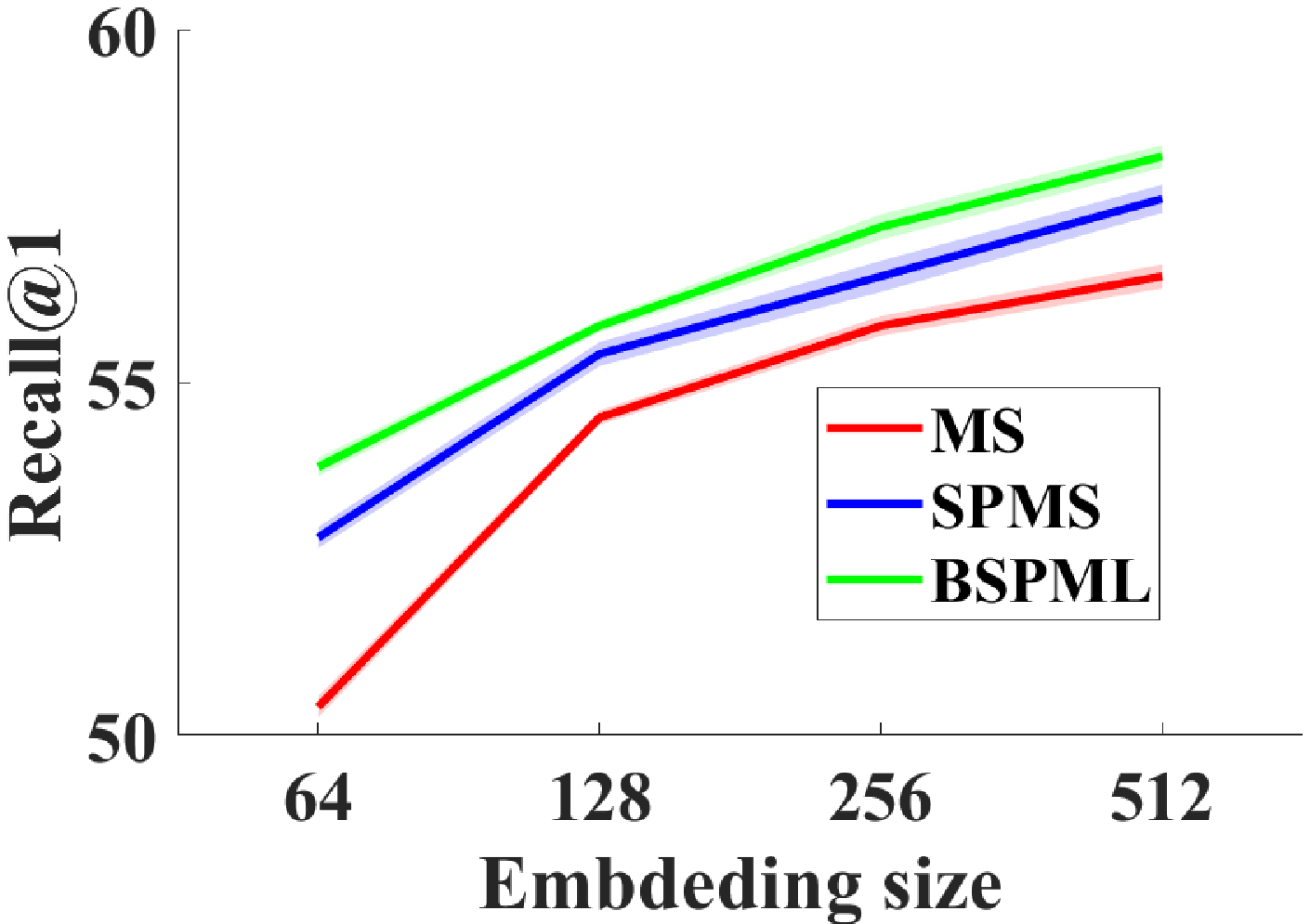}
	}
	\subfigure[In-shop]{
		\centering
		\includegraphics[width=2.0in]{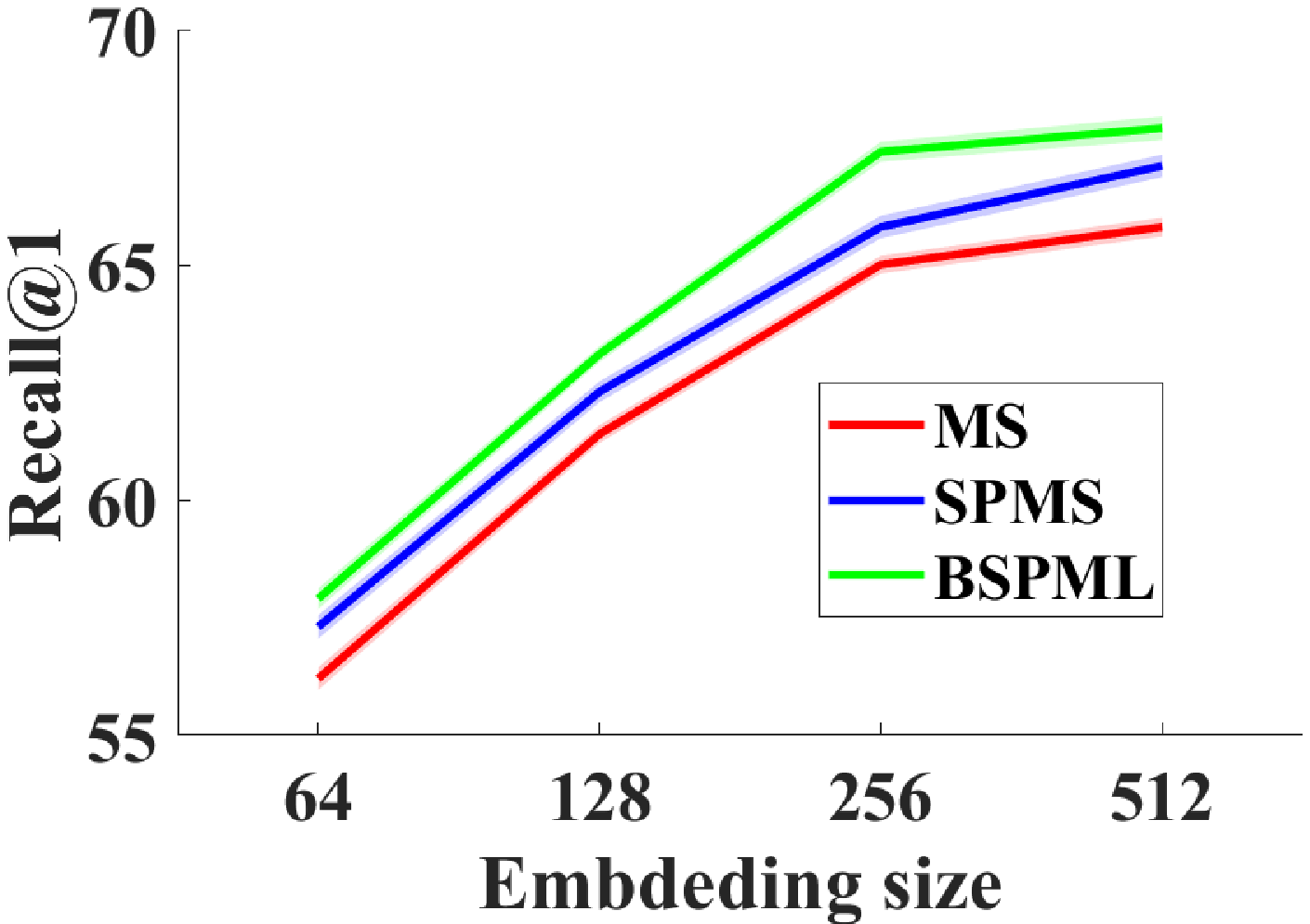}
	}
		\vspace{-4pt}
	\caption{Recall@$1$  results (\%) with different embedding sizes  on  data sets with 20\% noisy samples.} \label{dim}
		\vspace{-8pt}
\end{figure*}

\subsection{Results and discussion}
Table \ref{clean} presents Recall@$K$ and NMI performance on four standard data sets. Benefiting from the SPL strategy,  our BSPML avoids getting stuck into one bad local optimal solution and thus achieves better generalization ability than MS loss that is the non-SPL version of our BSPML. Meanwhile, compared with other state-of-the-art DML algorithms, our  BSPML also obtains better performance.

Table \ref{noise}  shows Recall@$1$  performance on three noisy data sets with different noise ratios (from 10\% to 30\%). These results show that our BSPML obviously achieves better performance than non-robust  DML algorithms. Compared with robust  DML algorithms, our BSPML also has sufficient advantages. Specifically,  while  SNR loss seems to be helpful against the noisy feature, it has limited robustness against the noisy label that is a much trickier challenge. Proxy-Anchor loss and RDML loss attempt to reduce the influence of noisy samples, but they are still sensitive to large noise ratios. Our BSPML  is able to exclude noisy samples from the training process directly and thus has better robustness. 

Fig. \ref{parameterLambda} and  Fig. \ref{parameterMu} show the effect of $\lambda$ and $\mu$ on the sample weight. Note that  MAW (\ref{EqMAWandEqSDAW}) represents the average weight of all samples and SDAW  (\ref{EqMAWandEqSDAW}) implies the balance degree between the average sample weights of classes. Fig.  \ref{parameterLambda} clearly reveals that with the increase of $\lambda$, our BSPML tends to assign samples larger weights and thus allows hard samples with larger losses to join the training. Obviously, this phenomenon is consistent with the classic SPL strategy. Moreover, as shown in Fig.  \ref{parameterMu},   the average sample weights of classes are balanced as much as possible when  $\mu$ is set to a large value. This phenomenon shows the critical role of our balance regularization term. 

Fig.  \ref{dim} shows the Recall@$1$ performance of ablation experiments with varying embedding sizes $\{64, 128, 256, 512\}$ on the data sets with 20$\%$ noisy samples, where the  SPMS  is generated by introducing SPL strategy into MS loss without the balance regularization term. Compared with the original MS loss, the performance of SPMS is improved regardless of the embedding size. However, SPMS still faces the challenge of the unbalanced average sample weights among classes. Benefiting from the balance regularization term, our BSPML achieves the best performance.

\section{Conclusion}
In this paper,  we build a connection between noisy samples and hard samples in the framework of self-paced learning and propose a \underline{B}alanced \underline{S}elf-\underline{P}aced \underline{M}etric \underline{L}earning (BSPML) algorithm with a novel denoising multi-similarity formulation to deal with noisy data in DML effectively.  
Specifically,  our BSPML algorithm treats noisy samples as extremely hard samples and adaptively excludes them from the model training by sample weighting.
Experimental results on several standard data sets demonstrate that our BSPML algorithm has better generalization ability and robustness than the state-of-the-art robust DML approaches. Importantly, we also theoretically prove the convergence for our proposed algorithms. 

\section{Acknowledgments}
Bin Gu was partially supported by the National Natural Science Foundation of China (No:61573191).

\bibliography{aaai23}

\end{document}